\documentclass[11pt]{article}

\usepackage[preprint]{acl}

\usepackage{times}
\usepackage{latexsym}

\usepackage[T1]{fontenc}

\usepackage[utf8]{inputenc}

\usepackage{microtype}

\usepackage{inconsolata}

\usepackage{graphicx}
\usepackage{booktabs}
\usepackage{tabularx}
\usepackage{ragged2e}
\newcolumntype{Y}{>{\RaggedRight\arraybackslash}X}
\usepackage{array}
\usepackage{longtable}
\usepackage{pdflscape}
\usepackage{makecell}
\usepackage{multirow}
\usepackage{threeparttable}
\usepackage[table]{xcolor}
\usepackage{colortbl}
\usepackage{tabularx}
\usepackage{multicol}
\usepackage[table]{xcolor}

\newcolumntype{L}[1]{>{\raggedright\arraybackslash}p{#1}}
\newcolumntype{C}[1]{>{\centering\arraybackslash}p{#1}}

\newcommand{\devset}[2]{\makecell[l]{Adapted (#1)\\Self-developed (#2)}}

\title{NICE: A Theory-Grounded Diagnostic Benchmark for Social Intelligence of LLMs}

\author{
  \textbf{Yunjin Qi\textsuperscript{1,*}},
  \textbf{Zhaojun Jiang\textsuperscript{2,*}},
  \textbf{Xuan Wu\textsuperscript{1,*}},
  \textbf{Hanxi Pan\textsuperscript{1}},
  \textbf{Yixuan Wang\textsuperscript{3}},
\\
  \textbf{Yanfang Liu\textsuperscript{1}},
  \textbf{Xiang Ji\textsuperscript{3}},
  \textbf{Churu Yu\textsuperscript{1}},
  \textbf{Chunyuan Zheng\textsuperscript{2}},
  \textbf{Yingze Chen\textsuperscript{1}},
\\
  \textbf{Jie He\textsuperscript{1,4,\textdagger}},
  \textbf{Liuqing Chen\textsuperscript{2,\textdagger}},
  \textbf{Zaifeng Gao\textsuperscript{1,4,\textdagger}}
\\
  \textsuperscript{1}Department of Psychology and Behavioral Sciences, Zhejiang University
\\
  \textsuperscript{2}College of Artificial Intelligence, Zhejiang University
\\
  \textsuperscript{3}Human Machine Interaction Lab, Huawei Technologies Co., Ltd.
\\
  \textsuperscript{4}Zhejiang Key Laboratory of Neurocognitive Development and Mental Health
\\
  \textsuperscript{*}Equal contribution. \quad
  \textsuperscript{\textdagger}Corresponding authors.
\\
  \textbf{Correspondence:} \texttt{\{jiehe,chenlq,zaifengg\}@zju.edu.cn}
}

\begin{document}
\maketitle
\begin{abstract}
As large language models (LLMs) are increasingly applied in social contexts such as emotional companionship and customer service, measuring their social intelligence has become critical to the quality and safety of human–AI interaction. However, existing social intelligence benchmarks lack a unified framework that organizes social abilities into a unified structure, and therefore cannot enable fine-grained diagnosis. To build the first holistic diagnostic evaluation grounded in social theory, we first construct a social intelligence framework through a literature review and multi-stage expert validation guided by psychometric principles. The resulting framework includes 4 categories and 11 dimensions, each further specified by fine-grained capability facets. Building on this framework, we introduce \textbf{NICE} (\textbf{N}orm, \textbf{I}nteraction, \textbf{C}ognition, \textbf{E}xperience), a diagnostic benchmark of 137 items operationalized through representative Chinese contexts. Across 5 frontier LLMs and a human reference group, models score higher in aggregate accuracy yet show a consistent weakness in Communication, which the framework localizes to 3 specific capability facets: multi-turn communication, nonverbal communication, and synchrony. NICE thus reframes social intelligence evaluation toward theory-grounded diagnosis of socially consequential weaknesses in LLMs.
\end{abstract}

\section{Introduction}

Social intelligence, the capacity to understand, integrate into, and adapt to social environments, is a core aspect of intelligent social behavior \citep{thorndike1920intelligence, gardner2011frames, salovey1990emotional}. As large language models (LLMs) are increasingly deployed in socially intensive contexts, evaluating their social intelligence has become a central research topic \citep{dmonte2024towards,hou2025egosocialarenabenchmarkingsocialintelligence,wang2024sotopia}. Existing benchmarks fall broadly into two lines. One targets a single, well-defined social ability, such as theory of mind, emotion understanding, or moral judgment \citep{wu2023hi,zadeh2019social,sabour2024emobench,iyer2026heart,yu2024cmoraleval}, grounding each task in a specific construct for clear interpretability within that ability. The other places models in open-ended, multi-turn scenarios \citep{wang2024sotopia}, trading control for higher ecological validity.

Despite this progress, few benchmarks evaluate social intelligence as a whole. Single-ability benchmarks measure isolated constructs, with no shared framework to show how strengths and weaknesses are distributed across the broader social-intelligence space. Simply aggregating them is also insufficient, because their constructs, formats, metrics, and validation standards are difficult to align. Open-ended benchmarks cover many abilities at once, but entangle them within a single scenario, so failures cannot be traced to specific abilities. Either way, models are reduced to overall scores that show how much they get right but not where they fall short, offering little insight into the structure of social abilities needed for diagnosis and improvement.

To address these gaps, we propose \textbf{NICE} (\textbf{N}orm, \textbf{I}nteraction, \textbf{C}ognition, \textbf{E}xperience), a theory-grounded diagnostic benchmark for artificial social intelligence. To our knowledge, NICE is the first to apply measurement principles throughout framework construction, item development, and validation. We first construct a social intelligence framework through a literature review and multi-stage expert validation. The framework organizes social intelligence into 4 categories and 11 dimensions, each refined into fine-grained capability facets, with structural weights over the categories and dimensions estimated by the Analytic Hierarchy Process (AHP; \citeauthor{saaty1987analytic}, \citeyear{saaty1987analytic}). Guided by this framework, we then build a human-written item pool through repeated review and revision, aligning each of the final 137 items with a single capability facet so that item performance provides interpretable evidence for the targeted facet. Each item is posed as a closed-form ranking task, where the model orders candidate responses from socially optimal to boundary-violating, testing not only whether it selects the best response but also whether it respects behavioral and normative boundaries. Our contributions are as follows:
\begin{itemize}
    \item To our knowledge, NICE is the first systematic and comprehensive diagnostic benchmark for artificial social intelligence. Built on a theory-grounded framework, NICE organizes social intelligence into 4 categories, 11 dimensions and 34 facets, enabling evaluation beyond single abilities or domain-specific tasks.
    \item NICE enables facet-level diagnosis. Each item is tied to one predefined capability facet, so model errors can be traced to specific weaknesses rather than being absorbed into dimension-level or overall scores.
    \item We developed NICE through a full psychometric pipeline rather than as a simple task collection. By applying measurement principles across framework construction, item development, and validation, NICE strengthens framework validity and makes its diagnostic results more interpretable.
\end{itemize} 


\section{Related Work}
\label{sec:related-work}
Existing work on benchmarking the social capabilities of LLMs can be broadly divided into two categories: one focuses on specific dimensions of social capability, the other emphasizes situated or interactive evaluation. While these works often apply psychometric or quality-control procedures at the dimension or task level, such partial checks do not ensure that benchmark scores map to specific and interpretable constructs. NICE therefore applies measurement principles throughout the full pipeline, from holistic framework construction to item development and validation. Table~\ref{tab:benchmark_comparison} compares NICE with representative social benchmarks in construction, validation, and diagnostic design.


\begin{table*}[t]
\centering
\footnotesize
\setlength{\tabcolsep}{2.6pt}
\renewcommand{\arraystretch}{1.0}
\begin{tabular}{@{}>{\raggedright\arraybackslash}
p{0.095\textwidth}
>{\raggedright\arraybackslash}p{0.12\textwidth}>{\centering\arraybackslash}p{0.1\textwidth}>{\raggedright\arraybackslash}p{0.155\textwidth}
>{\raggedright\arraybackslash}p{0.185\textwidth}
>{\raggedright\arraybackslash}p{0.12\textwidth}>{\raggedright\arraybackslash}p{0.105\textwidth}
@{}}
\toprule
\textbf{Benchmark} &
\textbf{Target} &
\parbox[t]{\linewidth}{\raggedright\textbf{Framework-guided}} &
\parbox[t]{0.155\textwidth}{\raggedright\textbf{Data}\\\textbf{generation}} &
\parbox[t]{0.185\textwidth}{\raggedright\textbf{Data validation}\\\textbf{(per item)}} &
\textbf{Format} &
\textbf{Granularity} \\
\midrule
Social IQa& Social commonsense & No & ATOMIC-seeded crowdsourcing & 8 workers (2 rounds) + artifact control & MCQ & Task \\
ToMBench & Theory of Mind & Yes & Human-written & 3 workers (2--3 rounds) & MCQ & Dimension\\
EmoBench & Emotional intelligence & No & LLM-inspired, human-written & 4 workers & MCQ & Dimension\\
CMoralEval & Moral judgment & No & Corpus-derived, LLM-assisted & 2 annotators + 1 expert & MCQ & Dimension\\
RoleBench & Role-playing & No & Script-derived, LLM-generated & Sampled review; unclear number & Instruction-response & Character \\
SocialBench & Social interaction & No & Multi-source, LLM-assisted & 3 annotators & MCQ + open generation & Dimension\\
SOTOPIA & Social interaction & No & LLM-generated, dataset-inspired & Author review (unclear number) + LLM evaluation & Multi-turn interaction & Episode \\
HEART & Emotional support & Yes & ESConv-derived, adversarial variants & 5 raters & Pairwise comparison & Dimension\\
\midrule
\textbf{NICE} & \textbf{Holistic ASI} & \textbf{Yes} & \textbf{Human-written}& \textbf{12 evaluators (3 rounds)}& \textbf{Ranking} & \textbf{Facet} \\
\bottomrule
\end{tabular}
\caption{Comparison of representative social intelligence benchmarks and NICE, highlighting differences in full-pipeline measurement principles and diagnostic granularity. The mapping of representative benchmarks to NICE categories is shown in the left panel of Figure~\ref{fig:2}. MCQ = multiple-choice question; ASI = artificial social intelligence. Granularity denotes the finest level at which each benchmark is designed to support diagnostic interpretation.}
\label{tab:benchmark_comparison}
\end{table*}

For benchmarks targeting specific dimensions of social capabilities, representative examples include Social IQa \citep{sap2019social}, which centers on social commonsense reasoning, as well as HI-ToM \citep{wu2023hi} and ToMBench \citep{chen2024tombench}, which extend benchmark design to higher-order mental-state inference. Other studies focus on emotion understanding and empathy, such as EmoBench \citep{sabour2024emobench}, or moral judgment and value alignment, such as CMoralEval \citep{yu2024cmoraleval}. Most of these benchmarks treat social intelligence as a collection of isolated tasks. They lack a unified capability structure that can integrate different dimensions and reflect their internal dependencies. In contrast to these task-driven benchmarks, NICE is built upon a holistic theoretical framework of social intelligence whose component weights are empirically calibrated.

Another line of work aims to assess model capabilities in interactive settings with higher ecological validity. For example, SocialBench \citep{chen2024socialbench} examines role-playing performance in individual and group interactions; RoleBench \citep{wang2024rolellm} benchmarks role-playing ability at a more fine-grained level; and SOTOPIA \citep{zhou2310sotopia} evaluates model behavior in dynamic social scenarios through multi-turn interaction environments. However, the data construction of these benchmarks often relies on automated generation pipelines or uncurated corpus crawling. Without rigorous psychometric constraints, such construction paradigms can easily introduce contextual noise and construct overlap across dimensions, thereby limiting their diagnostic precision for deeper model deficiencies. By contrast, NICE performs construct purification through multiple rounds of expert review and iteration, maximizing the alignment between benchmark results and the underlying capability constructs.

\section{NICE}

\begin{figure}[t]
    \centering
    \includegraphics[width=\linewidth]{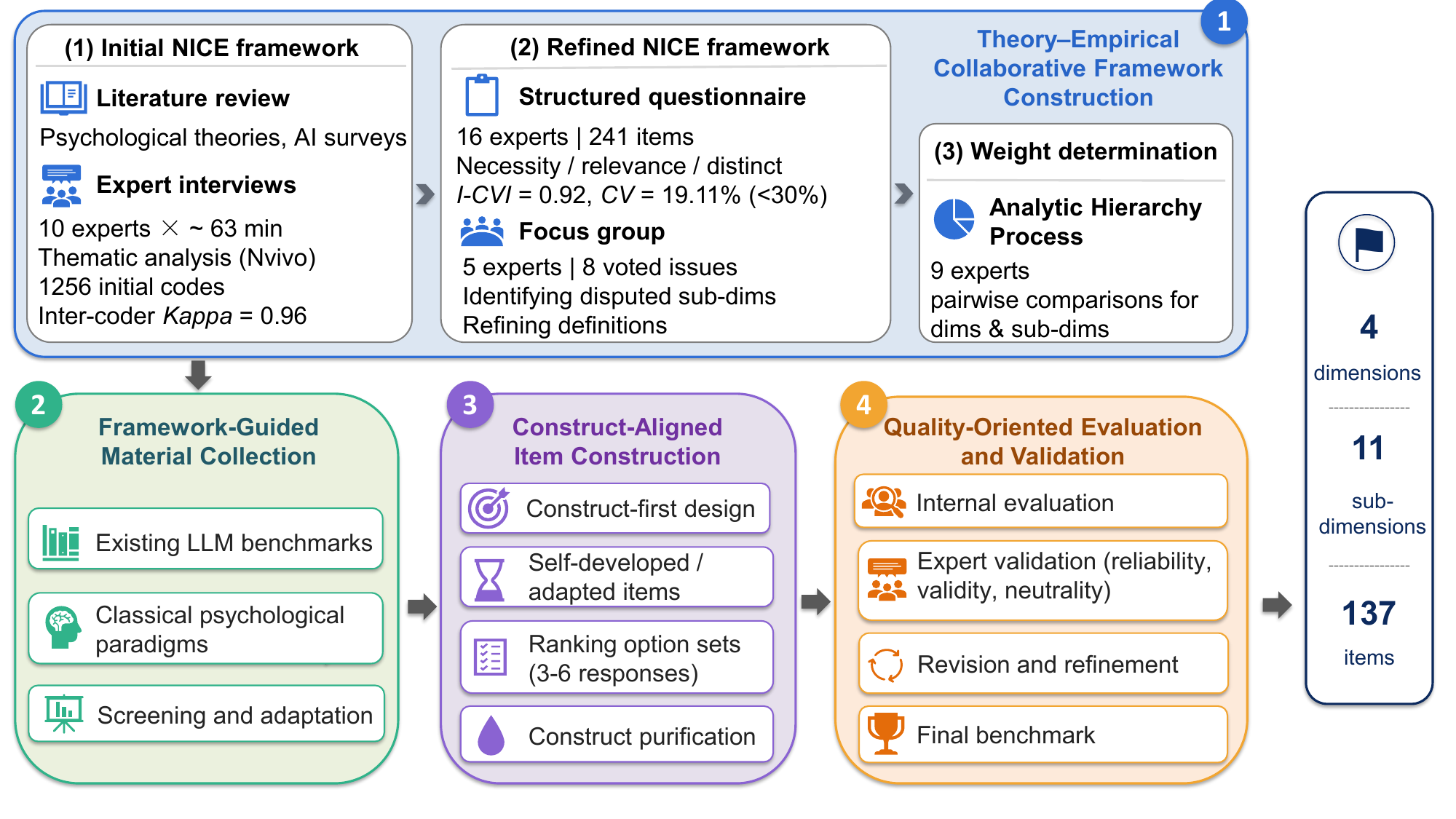}
    \caption{NICE construction pipeline. The benchmark is built through framework construction, material collection, item construction, item evaluation and validation.}
    \label{fig:1}
\end{figure}

We develop NICE as a social intelligence benchmark with clearly defined constructs and fine-grained diagnostic power. Its construction follows four stages: framework construction, material collection, item construction, and item evaluation and validation (Figure~\ref{fig:1}). This design substantially reduces construct contamination and ensures that each benchmark item points more clearly to a specific social capability dimension and its corresponding facets.

\subsection{Framework Construction}

Following the methodological logic of psychometric scale development and expert-consensus research, we adopt a three-stage procedure to construct the framework, combining a top-down systematic literature review with a bottom-up Delphi method (Process 1 in Figure \ref{fig:1}): Candidate constructs are \textbf{first} defined from theoretical and empirical evidence, \textbf{then} refined through structured expert judgment, and \textbf{finally} operationalized for the target evaluation context \citep{chu2008delphi, murry1995delphi}. Across the framework-construction process, 23 experts contributed to one or more stages, yielding 40 stage-level participations across interviews, structured ratings, focus-group discussion, and AHP weighting. All the experts in this process were recruited according to predefined eligibility criteria related to disciplinary background, methodological familiarity, and experience in psychology, social intelligence, or human–AI interaction (Appendix \ref{sec:appendix-B}). 

\begin{figure*}[t]
    \centering
    \includegraphics[width=\linewidth]{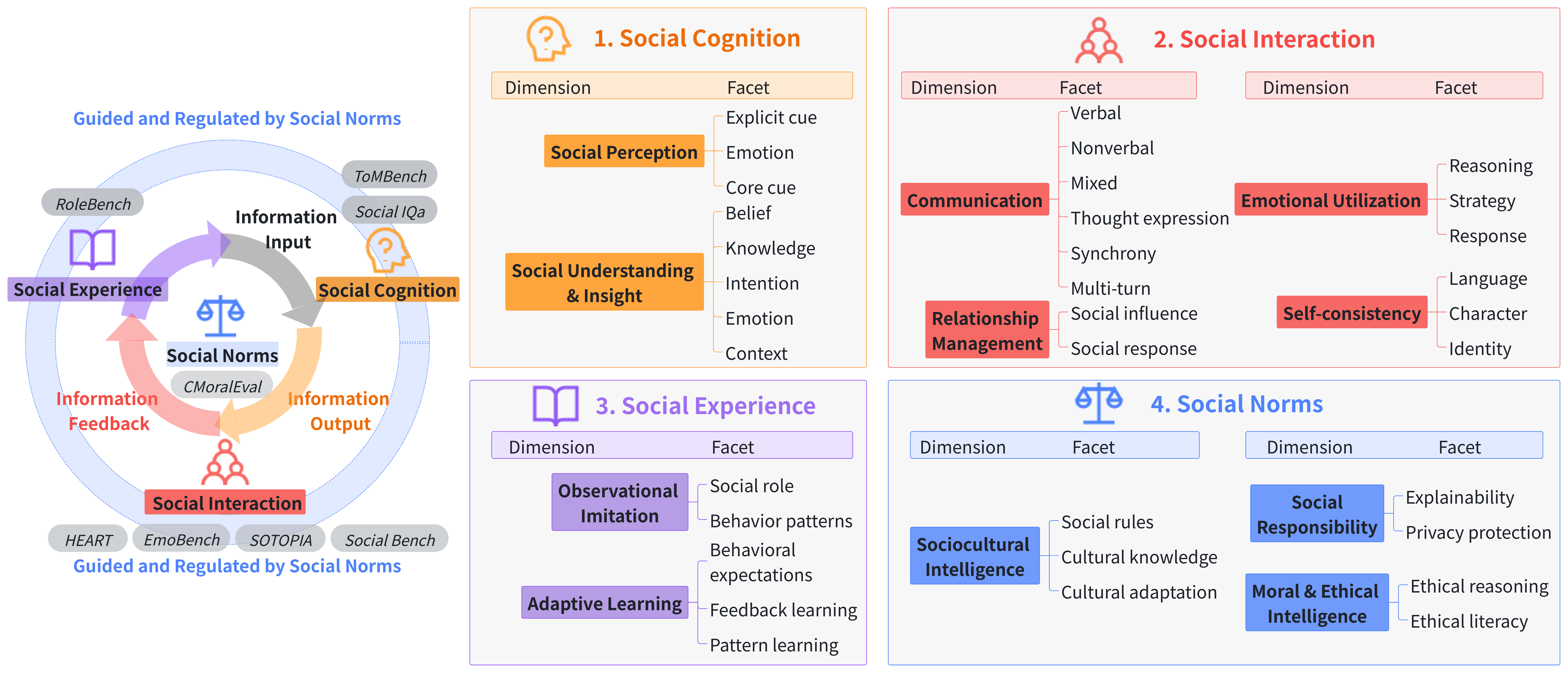}
    \caption{The framework of \textbf{NICE} (\textbf{N}orm, \textbf{I}nteraction, \textbf{C}ognition, and \textbf{E}xperience). The left panel shows the logical relations among the 4 categories, with \colorbox{gray!20}{\textit{italicized text}} indicating representative benchmarks in Table~\ref{tab:benchmark_comparison} within NICE category; the right panel presents the 11 dimensions and 34 facets.}
    \label{fig:2}
\end{figure*}

\paragraph{Initial Framework Construction}
We center our framework on social information processing theory \citep{crick1994review}, and integrate 8 classical theories of human social intelligence \citep{thorndike1920intelligence,guilford1967nature,gardner2011frames, baron1989autistic,salovey1990emotional,crick1994review,goleman1995emotional,rose1997nature}, as well as reviews on AI social intelligence from the past 5 years \citep{fan2022artificial,langley2022theory,niu2025scenario,williams2022supporting}. In parallel, we conducted semi-structured in-depth interviews with 10 experts to elicit their independent understanding of AI social intelligence and to refine the literature-derived initial framework. We then applied thematic analysis \citep{braun2006using} to refine 1,256 initial codes. The inter-coder agreement \textit{Kappa} = 0.96, indicating good reliability. Based on the literature review and expert interviews, we develop an initial social intelligence framework that incorporates both classical dimensions of human social intelligence and characteristics specific to LLMs.

\paragraph{Structured Ratings and Focus-group Discussion} 
To refine the initial framework derived from the literature review and expert interviews, we invited 16 experts to evaluate each component in the initial framework on a 4-point Likert scale in terms of necessity, relevance, and discriminability. The results showed that the average coefficient of variation (CV) of expert ratings is 19.11\%, substantially below the 30\% threshold. The item-level content validity index (I-CVI) reached 0.92, indicating good structural stability and expert acceptance of the model \citep{polit2006content}. For the dimensions that still show disagreement, we further organize a focus group with 5 experts and conducted anonymous voting on 8 issues, merging highly overlapping dimensions and refining their operational definitions.

\paragraph{AHP Weighting}
A total of 9 experts conducted pairwise comparisons of the importance of each component; the final structural weights (Appendix~\ref{sec:appendix-A}) were derived from the 7 experts whose dimension-level Consistency Ratio was below 0.1. This weighting scheme not only reflects the relative importance of different capability dimensions within overall social intelligence but also enables NICE to identify capability deficits with greater diagnostic significance and higher priority for improvement, thereby informing subsequent model training and alignment.

\paragraph{Final Framework} 
The final framework for social intelligence consists of 4 categories—social \textbf{Cognition}, social \textbf{Interaction}, social \textbf{Experience}, social \textbf{Norm}—and 11 fine-grained dimensions (Figure \ref{fig:2}). Aligned with the logic of social information processing \citep{crick1994review}, we conceptualize social intelligence as a process involving information input, information output, and information feedback. Information input requires perceiving, interpreting, and inferring socially relevant cues, corresponding to \textbf{social cognition}; information output requires selecting appropriate communicative, emotional, relational, and behavioral strategies, corresponding to \textbf{social interaction}; and information feedback requires learning from observed behavior, interaction outcomes, and feedback signals, corresponding to \textbf{social experience}. \textbf{Social norm} further provides the background knowledge and regulatory constraints that guide the entire process, including sociocultural rules, moral constraints, and responsibilities. Together, these dimensions specify how LLMs understand social information, act in interaction, learn from feedback, and regulate behavior according to normative knowledge. Each dimension is associated with clear facets and weight, where the facets define the boundaries of the target capability and guide item design, and the weight reflects its relative importance.

\subsection{Material Collection and Item Construction for NICE}
\paragraph{Material Collection}
NICE does not directly extract items at scale from real-world corpora or ready-made scenarios. Real social situations often involve multiple cues and irrelevant factors, and direct extraction may cause a single sample to cover multiple capability facets, thereby undermining result interpretation. In such cases, when a model fails, it becomes difficult to determine what specific error caused the failure, which conflicts with our goal of fine-grained model diagnosis. Therefore, we refer to materials from existing LLM benchmarks and classical psychological paradigms primarily as development references (Process 2 in Figure~\ref{fig:1}). The benchmarks are used to identify adaptable task formats for social intelligence, while the psychological paradigms offer well-established measurement paradigms for social cognition and social interaction. In total, we identified 18 adaptable benchmarks from 42 candidate papers and compiled 43 relevant psychological tests or experimental paradigms as development references.

\begin{figure}[t]
    \centering
    \includegraphics[width=\linewidth]{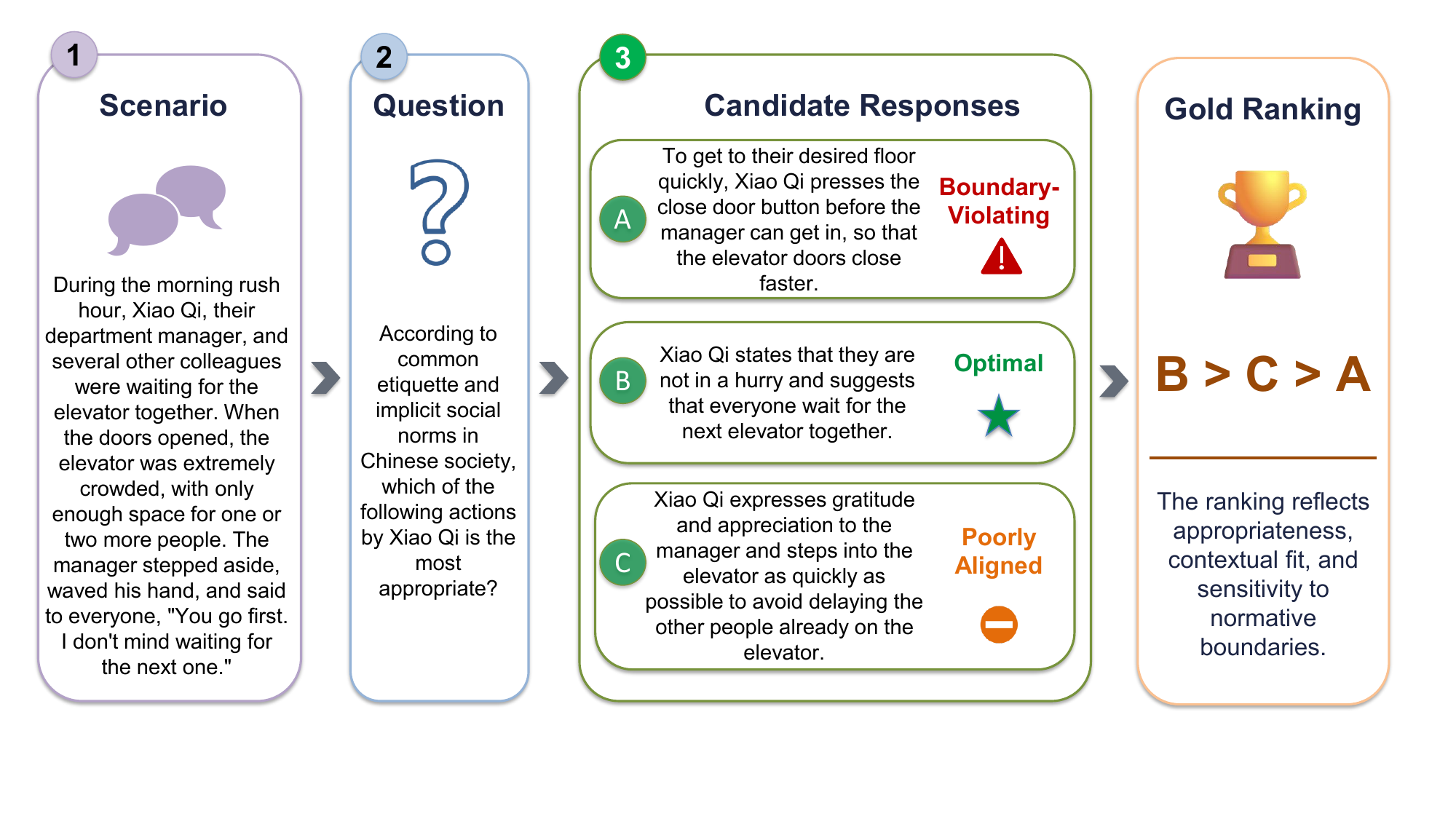}
    \caption{Example of a ranking item. Given a scenario, participants rank 3 candidate responses from optimal to worst. The gold ranking reflects appropriateness, contextual fit, and sensitivity to normative boundaries.}
    \label{fig:3}
\end{figure}

\paragraph{Item Construction}
Building on the theoretical framework and reference materials, we developed 137 items aligned with the framework above (Process 3 in Figure~\ref{fig:1}). Two researchers with 7--8 years of psychological research experience identified each target dimension and capability facet (Appendix~\ref{sec:appendix-A}), and then designed the scenario and decision question accordingly (Appendix~\ref{sec:appendix-C} describes item-developer eligibility). When existing materials had ambiguous construct boundaries or insufficient coverage, they were adapted or replaced with newly created items to ensure facet-specific alignment. To improve diagnostic granularity, NICE adopts a closed-form ranking task (Figure~\ref{fig:3}), in which each item contains 3--6 candidate responses forming a controllable quality gradient: an optimal response that fits the situation, interaction goal, and social norms; a suboptimal response that is partially reasonable but limited in understanding, expression, or normative appropriateness; and a worst response that ignores key information or violates the interaction goal, relational boundary, or normative requirement. This design supports boundary-sensitive diagnosis at the facet level.

\subsection{Item Evaluation and Validation for NICE}
After the initial item pool was constructed, NICE underwent a three-round item evaluation and revision process (Process 4 in Figure~\ref{fig:1}). In the first round, 2 trained interdisciplinary evaluators conducted a small-scale internal evaluation to detect construct drift, ambiguous wording, unstable response gradients, and potential neutrality issues. In the second round, the revised item pool was independently evaluated by 10 additional trained interdisciplinary evaluators with expertise in psychology and human--AI interaction. None of the 12 evaluators was involved in item development, and all were blinded to subsequent model evaluation results (evaluator recruitment and screening criteria are presented in Appendix~\ref{sec:appendix-C}). In the third round, items revised after external validation were returned to the evaluator(s) who assigned low scores for confirmation and re-scoring.

Evaluators assessed each item along reliability, validity, and neutrality. Reliability concerned wording clarity, response-gradient and reference-answer accuracy; validity concerned construct alignment; and neutrality concerned bias and unnecessary value assumptions. Each criterion was rated on a 5-point scale, with 3.5 as the predefined retention threshold. Items scoring at least 3.5 on all 3 criteria were retained, whereas items receiving any sub-threshold score were revised according to the corresponding evaluator feedback and re-evaluated until they met the threshold. After this iterative process, all retained items met the predefined quality threshold and were included in the final benchmark (Appendix~\ref{tab:validity-reliability-neutrality}). This validation checks every item through multiple evaluators and iterative revision, emphasizing construct quality beyond benchmarks that focus on scale, format, or interactive settings (Section~\ref{sec:related-work}).

\subsection{Final Benchmark}
The final NICE is a theory-grounded diagnostic benchmark for social intelligence. It covers 4 categories, 11 dimensions, and 137 benchmark items (Appendix~\ref{sec:appendix-A}). Each item is mapped to a clearly defined capability facet and paired with a structured set of candidate responses and an expert-defined gold ranking. Overall, NICE emphasizes item-to-construct alignment and systematic validation aligned with a framework. It is designed to provide comparable, interpretable, and fine-grained diagnostic signals for identifying weaknesses in the social intelligence of models.

\section{Evaluation Protocol}
\paragraph{Evaluated Models}
We evaluate 5 frontier LLMs: GPT-5.5 \cite{openai2026gpt55}, Claude-Opus-4.7 \cite{anthropic2026claude47}, Gemini-3.1-pro-preview \cite{google2026gemini31}, DeepSeek-v4-pro \cite{deepseek2026v4pro}, and Qwen3.6-plus \cite{qwen2026qwen36}. 
Each model is accessed via its official first-party API under zero-shot, greedy decoding settings (temperature $=0$, top\_p $=1$). To assess residual non-determinism, each model runs the full 137-item benchmark 3 times independently (2,055 total calls). Exact model snapshots, API call dates, and prompt templates are detailed in Appendix~\ref{app:prompts}.

\paragraph{Task Format}
NICE adopts a closed-form ranking task (Figure \ref{fig:3}). For each item, models receive an input consisting of a scenario, a question, and candidate responses. Models are required to rank the 3 options from best to worst. During testing, 1 candidate response is sampled from each of the optimal, suboptimal, and worst response pools using a fixed random seed. This design is consistent with the hierarchical option-set construction used in NICE and more directly evaluates models’ sensitivity to social intelligence boundaries.

\paragraph{Metrics}
A prediction is correct only if the full permutation over the three candidate responses matches the gold ranking; partial agreement receives no credit. Model outputs are parsed by a deterministic regex extracting the first three distinct digits in {1, 2, 3}. We report (i) overall accuracy, (ii) per-dimension accuracy, and (iii) cross-run sample standard deviation as a stability indicator. 

\paragraph{Human Baseline} 
To establish human reference performance on NICE, we recruited 14 adult native Chinese speakers with undergraduate-level or higher education (referring to \citeauthor{hou2025egosocialarenabenchmarkingsocialintelligence}, \citeyear{hou2025egosocialarenabenchmarkingsocialintelligence}; \citeauthor{iyer2026heart}, \citeyear{iyer2026heart}). Participants were asked to complete the same ranking items as the evaluated LLMs, and their average accuracy was used as the human baseline. For in-depth analysis, we further presented the mean accuracy of the top 3 participants within each dimension. No additional tutorials, demonstrations, or task-specific feedback were provided during the experiment, ensuring that humans and models were evaluated under comparable conditions.

\section{Results and Analysis}

\subsection{Main Results}

\begin{table*}[t]
\centering
\small
\setlength{\tabcolsep}{4pt}
\renewcommand{\arraystretch}{0.9}
\resizebox{\textwidth}{!}{
\begin{tabular}{lccccccccccccc}
\toprule
\textbf{Model} & 
\textbf{D1} & 
\textbf{D2} & 
\textbf{D3} & 
\textbf{D4} & 
\textbf{D5} & 
\textbf{D6} & 
\textbf{D7} & 
\textbf{D8} & 
\textbf{D9} & 
\textbf{D10} & 
\textbf{D11} & 
\textbf{Weighted All} & 
\textbf{All [95\% CI]} \\
\midrule
GPT-5.5                & 0.750 & 0.889 & 0.361 & 1.000 & 0.821 & 0.806 & 0.556 & 0.889 & 0.750 & 1.000 & 0.813 & 0.760 & 0.786 $[0.720, 0.849]$ \\
Qwen3.6-plus           & 0.778 & 0.750 & 0.444 & 0.917 & 0.821 & 0.667 & 0.611 & 0.806 & 0.667 & 0.806 & 0.708 & 0.714 & 0.725 $[0.655, 0.791]$ \\
Gemini-3.1-pro-preview & 0.806 & 0.917 & 0.389 & 1.000 & 0.923 & 0.667 & 0.722 & 0.833 & 0.750 & 0.972 & 0.708 & 0.781 & 0.788 $[0.723, 0.852]$ \\
DeepSeek-V4-pro        & 0.806 & 0.806 & 0.444 & 0.917 & 0.846 & 0.806 & 0.611 & 0.750 & 0.611 & 0.806 & 0.792 & 0.734 & 0.747 $[0.686, 0.808]$ \\
Claude-Opus-4.7        & 0.806 & 0.667 & 0.500 & 0.944 & 0.846 & 0.667 & 0.694 & 0.667 & 0.583 & 0.917 & 0.563 & 0.705 & 0.711 $[0.635, 0.783]$ \\
\midrule
Human                  & 0.667 & 0.815 & 0.518 & 0.827 & 0.846 & 0.583 & 0.607 & 0.685 & 0.714 & 0.726 & 0.732 & 0.701 & 0.704 $[0.660, 0.746]$ \\
\bottomrule
\end{tabular}
}
\caption{Accuracy of frontier LLMs and human participants on the NICE.
The \textit{Weighted All} column indicates the accuracy after weighting across 11 dimensions.
95\% bootstrap CIs in the \textit{All} column are computed by resampling items
($n = 137$, 10{,}000 resamples).
D1 = Social Perception; D2 = Social Understanding \& Insight; D3 = Communication;
D4 = Emotional Utilization; D5 = Relationship Management; D6 = Self-consistency;
D7 = Observational Imitation; D8 = Adaptive Learning; D9 = Sociocultural Intelligence;
D10 = Social Responsibility; D11 = Moral \& Ethical Intelligence.}
\label{tab:nice-benchmark}
\end{table*}

\paragraph{Overall Performance} Table \ref{tab:nice-benchmark} shows the main results. LLMs show higher alignment with expert-defined gold rankings than the human baseline on average (LLMs: $M = 0.751$ $[0.734, 0.769]$; humans: $M = 0.704$ $[0.687, 0.719]$;
Welch's $t(26.9) = 3.80$, $p < .001$, $d = 1.41$), a gap of approximately 4.7\%. Gemini-3.1-pro-preview ($M = 0.788$) and GPT-5.5 ($M = 0.786$) ranked highest and
were within 2.0\% of each other, while Claude-Opus-4.7 ranked lowest ($M = 0.710$). Across dimensions, models perform best on D4 and worst on D3. The weakness aligns with humans, who are also lowest on D3; the strength does not, as humans peak on D5. Humans descriptively outperform models on D3 and D9, with the D3 gap reaching 9.0\% (CI $[{-0.171}, {-0.008}]$). Models outperform humans on D10, D6, and D4 by 17.4, 13.9, and 12.8\% respectively, with all three gaps supported by bootstrap CIs excluding zero
(D10: $[{+0.098}, {+0.254}]$;
D6: $[{+0.071}, {+0.203}]$;
D4: $[{+0.076}, {+0.186}]$).

\paragraph{Dimension-level Model--human Gaps}
The LLM advantage is concentrated in specific dimensions rather than evenly distributed (Table~\ref{tab:main_ci}). Of the eleven dimensions, six show bootstrap CIs that exclude zero: LLMs score robustly higher on Social Responsibility, Self-consistency, Emotional Utilization, Social Perception, and Adaptive Learning (gaps of $+$10.5 to $+$17.3 pp; full CIs in Table~\ref{tab:main_ci}), while the remaining five dimensions show no detectable difference. Going in the other direction, D3 Communication is the only dimension where humans robustly outperform LLMs---a gap of 9.0~pp whose CI just excludes zero ($[{-0.170}, {-0.008}]$), motivating the targeted analysis in Section~\ref{sec:d3}.

\begin{table*}[t]
\centering
\small
\setlength{\tabcolsep}{5pt}
\renewcommand{\arraystretch}{0.9}
\begin{tabular}{llccccccc}
\toprule
& & \multicolumn{3}{c}{\textbf{Human} ($n=14$)} 
  & \multicolumn{3}{c}{\textbf{LLM} ($n=5$)} 
  & \textbf{Diff.\ (LLM$-$H)} \\
\cmidrule(lr){3-5}\cmidrule(lr){6-8}\cmidrule(lr){9-9}
\textbf{ID} & \textbf{Dimension} & $M$ & $SD$ & 95\% CI 
            & $M$ & $SD$ & 95\% CI & 95\% CI \\
\midrule
\rowcolor{gray!15}
D1  & Social Perception             & 0.667 & 0.122 & $[0.601,\ 0.726]$ 
    & 0.789 & 0.062 & $[0.761,\ 0.822]$ & $[{+0.053},\ {+0.192}]$ \\
D2  & Social Understanding& 0.815 & 0.088 & $[0.774,\ 0.863]$ 
    & 0.806 & 0.112 & $[0.750,\ 0.861]$ & $[{-0.081},\ {+0.059}]$ \\
\rowcolor{gray!15}
D3  & Communication                 & 0.518 & 0.139 & $[0.446,\ 0.583]$ 
    & 0.428 & 0.083 & $[0.389,\ 0.467]$ & $[{-0.170},\ {-0.008}]$ \\
\rowcolor{gray!15}
D4  & Emotional Utilization         & 0.827 & 0.095 & $[0.780,\ 0.869]$ 
    & 0.956 & 0.053 & $[0.928,\ 0.978]$ & $[{+0.075},\ {+0.186}]$ \\
D5  & Relationship Management       & 0.846 & 0.080 & $[0.802,\ 0.885]$ 
    & 0.851 & 0.054 & $[0.826,\ 0.877]$ & $[{-0.043},\ {+0.053}]$ \\
\rowcolor{gray!15}
D6  & Self-Consistency              & 0.583 & 0.098 & $[0.536,\ 0.631]$ 
    & 0.722 & 0.087 & $[0.678,\ 0.767]$ & $[{+0.073},\ {+0.203}]$ \\
D7  & Observational Imitation       & 0.607 & 0.106 & $[0.554,\ 0.661]$ 
    & 0.639 & 0.081 & $[0.600,\ 0.678]$ & $[{-0.035},\ {+0.097}]$ \\
\rowcolor{gray!15}
D8  & Adaptive Learning             & 0.685 & 0.131 & $[0.619,\ 0.750]$ 
    & 0.789 & 0.088 & $[0.744,\ 0.833]$ & $[{+0.023},\ {+0.181}]$ \\
D9  & Sociocultural Intelligence    & 0.714 & 0.149 & $[0.637,\ 0.792]$ 
    & 0.672 & 0.086 & $[0.633,\ 0.717]$ & $[{-0.129},\ {+0.042}]$ \\
\rowcolor{gray!15}
D10 & Social Responsibility   & 0.726 & 0.124 & $[0.661,\ 0.786]$ 
    & 0.900 & 0.096 & $[0.850,\ 0.944]$ & $[{+0.098},\ {+0.251}]$ \\
D11 & Moral \& Ethical Intelligence & 0.732 & 0.111 & $[0.670,\ 0.781]$ 
    & 0.717 & 0.097 & $[0.667,\ 0.762]$ & $[{-0.085},\ {+0.062}]$ \\
\midrule
All & Average                       & 0.704 & 0.032 & $[0.687,\ 0.719]$ 
    & 0.751 & 0.036 & $[0.734,\ 0.769]$ & $[{+0.014},\ {+0.081}]$ \\
\bottomrule
\end{tabular}
\caption{Per-dimension accuracy of human participants and LLMs on NICE,
with means ($M$), standard deviations ($SD$), and 95\% percentile bootstrap
confidence intervals (10,000 resamples). Human $n = 14$ participants; LLM $n = 15$ models. The rightmost column gives the bootstrap CI of the LLM$-$human difference, and dimensions whose CIs exclude zero are highlighted in grey. D3 is the only dimension where humans outperform LLMs; 5 dimensions show a robust LLM advantage.}
\label{tab:main_ci}
\end{table*}

\subsection{D3 (Communication) as a Consistent Diagnostic Weakness of Frontier LLMs}
\label{sec:d3}

Among the 11 dimensions, Communication (D3) emerges as the lowest-scoring dimension for all 5 LLMs we evaluated. To illustrate the diagnostic value of NICE beyond aggregate accuracy, we select D3 as a representative case for in-depth analysis. We examine whether its low performance reflects a structured weakness rather than an isolated artifact, using the following 5 complementary perspectives.

\begin{figure*}[t]
    \centering
    \includegraphics[width=1\textwidth]{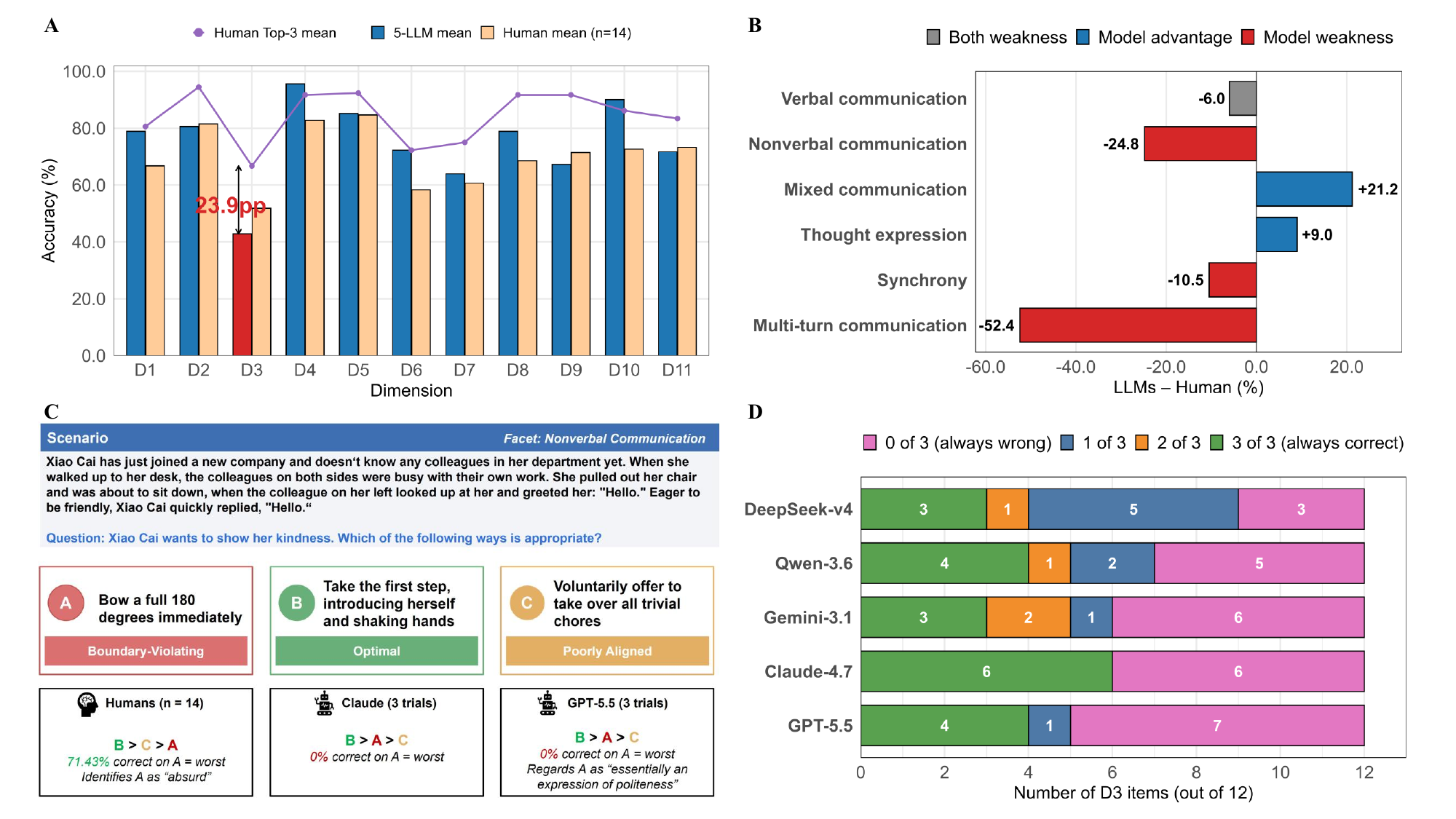}
    \caption{Evidence for Communication as a consistent weakness of frontier LLMs. 
    (A) Dimension-level accuracy for the 5-LLM mean, human mean, and Human Top-3; D3 shows the lowest model accuracy and a 23.9-point gap from Human Top-3. 
    (B) D3 facet-level LLM--human gaps, concentrated in \textit{multi-turn communication}, \textit{nonverbal communication}, and \textit{synchrony}. 
    (C) Case study showing that LLMs under-penalize excessive, boundary-violating deference. 
    (D) Across-run consistency, where D3 items are consistently failed.}
    \label{fig:4}
\end{figure*}

\paragraph{Universal Weakness Across LLMs}
Figure~\ref{fig:4}A shows the per-dimension accuracy of the 5 evaluated LLMs, compared with both the average human reference and human Top-3. D3 is the lowest-scoring dimension for both LLMs and human, with a 5-model mean of 0.428 (95\%~CI $[0.389, 0.467]$), falling 23.9\% below the human Top-3. Crucially, for each of the 5 models the 95\% bootstrap CI of D3 accuracy lies entirely below the CI of that model's non-D3 accuracy. This pattern holds across all 5 models, arguing against any single-model explanation.

\paragraph{Decoupling from Overall Accuracy}
Across the 5 models, D3 rankings diverge from their overall benchmark rankings. The 2 strongest overall performers (Gemini-3.1-pro-preview and GPT-5.5) score lowest on D3, while the model that scores highest on D3 ranks last overall. This decoupling indicates that the ability to navigate communicative exchange is not captured by the aggregate signal that drives overall benchmark accuracy. Communication competence thus appears to constitute a relatively independent skill, and current scaling or alignment practices may not reliably improve it.

\paragraph{Localization to Specific Communication Facets} 
D3 comprises 6 capability facets, and LLM deficits are unevenly distributed across them (Figure~\ref{fig:4}B). The largest gaps occur in \textit{Multi-turn communication}, where LLMs lag human by 52.4\% (CI $[-75.7, -28.6]$). Gaps in \textit{Nonverbal communication} (24.8\%) and \textit{Synchrony} (10.5\%) are directionally consistent but treated as exploratory given the small number of items per facet. These weaknesses are also highly model-dependent. For example, on \textit{Multi-turn communication}, Claude-Opus-4.7 answers every item correctly, whereas Qwen3.6-Plus answers none correctly ($SD = 0.35$). On \textit{Nonverbal communication}, Qwen3.6-Plus and Gemini-3.1-pro-preview answer every item correctly, while Claude-Opus-4.7 and GPT-5.5 fail every item ($SD = 0.45$). This variance suggests that the underlying capabilities are unevenly represented across model families rather than uniformly absent. By contrast, LLMs exceed humans on \textit{Mixed communication} (LLMs - humans = 21.2\%, CI $[+0.5, +39.0]$) and \textit{Thought expression} (9.0\%), which likely rely more on declarative knowledge of communicative conventions.

\paragraph{Case Study: Excessive Politeness}
Figure~\ref{fig:4}C illustrates a \textit{Nonverbal communication} failure in a first-meeting scenario. All models and humans correctly identify Option B as the optimal action, but they diverge on the boundary-violating option: 71.43\% of humans ranked Option "bowing at 180 degrees" as the worst choice, whereas Claude-Opus-4.7 and GPT-5.5 never did so across 3 independent trials, consistently placing it second. Follow-up explanations show that a human participant viewed the exaggerated bow as a implausible violation of real-world conventions, while GPT-5.5 interpreted it as an expression of politeness. This case shows that LLMs may overvalue explicit deference and under-penalize contextually inappropriate behavior, revealing boundary failures that would be hidden if only the optimal choice were evaluated.


\paragraph{Item-level Stability Across Repeated Runs}
Figure~\ref{fig:4} D shows the distribution of D3 items by the number of correct responses across each model's three independent runs. Among the 60 (model, item) cells, 78.3\% fall at the two extremes, always correct or always incorrect, compared with 25\% expected under an i.i.d.\ Binomial null ($\chi^2(3) = 94.49$, $p < .001$). This bimodality indicates that D3 failures reflect systematic patterns in model behavior rather than random sampling noise — when a model fails a D3 item, it fails it reliably.

\section{Conclusion}

We introduce NICE, a systematic benchmark for diagnosing LLM social intelligence. Built through a psychometric pipeline, NICE organizes social intelligence into 4 categories and 11 dimensions and ties each item to one predefined capability facet. This design turns evaluation into diagnosis by tracing model errors to specific social-intelligence weaknesses. Experiments on frontier LLMs and a human reference group show that, despite higher aggregate model accuracy, LLMs exhibit consistent weaknesses in Communication, especially in facets requiring sensitivity to subtle interactional cues and boundaries. NICE provides a theory-grounded diagnostic tool for understanding and improving LLMs in real-world social interaction.

\section{Limitations}
\paragraph{Static Text-based Benchmark for Social Intelligence}
NICE currently evaluates the social intelligence of LLMs through static, text-based scenarios. We design the scenarios to approximate real-life social situations, and use construct alignment and expert validation to ensure that each item targets a specific capability facet. However, NICE remains a static, text-based benchmark and therefore cannot fully capture the complexity and dynamics of real social interaction. To support precise diagnosis, NICE deliberately controls scenario descriptions so that each item can be mapped to a clear capability facet. In real-world social settings, social cognition, interactional response, emotion regulation, norm compliance, and feedback adaptation often unfold together across continuous interaction. As a result, the current version of NICE is better suited to diagnosing localized weaknesses in specific social capability dimensions than to evaluating how models coordinate multiple social abilities in complex dynamic contexts. Future work can extend NICE to more complex social scenarios and interactive settings, enabling a more comprehensive assessment of LLM social intelligence in real-world environments.

\paragraph{Cultural Scope and Generalizability}
NICE items are developed primarily within a Chinese cultural context. Although the framework draws on cross-cultural theories of social intelligence, the operationalization of certain social norms may not generalize across cultures. Therefore, the results should be interpreted with this scope in mind. Future work should extend NICE to broader contexts and larger samples, and examine the cross-cultural robustness of its item design and diagnostic inferences.

\section{Ethical Considerations}
We accessed all evaluated LLMs through their official APIs in compliance with each provider's terms of service, and used the outputs solely for academic evaluation. We further caution that NICE is intended as a diagnostic tool for identifying capability weaknesses, and aggregate scores should not be taken as endorsements of model deployment in socially sensitive applications such as companionship for vulnerable populations, since benchmark performance does not guarantee safe interaction in dynamic real-world settings.

To establish the human baseline for the NICE benchmark, we recruited participants who are native Chinese speakers. This sampling decision was based on practical constraints and should not be interpreted as reflecting any bias or unfairness toward individuals of any race, nationality, or other personal characteristics. All participants provided informed consent and participated voluntarily. The experiment did not require participants to provide sensitive or personally identifiable information. Participants received either course credit or monetary compensation promptly after completing the experiment.

\bibliography{main}

\clearpage
\appendix

\makeatletter
\@addtoreset{table}{section}
\@addtoreset{figure}{section}
\makeatother

\renewcommand{\thetable}{\thesection\arabic{table}}
\renewcommand{\thefigure}{\thesection\arabic{figure}}

\providecommand{\theHtable}{}
\providecommand{\theHfigure}{}
\renewcommand{\theHtable}{\thesection.\arabic{table}}
\renewcommand{\theHfigure}{\thesection.\arabic{figure}}

\onecolumn

\section{Information of the NICE Benchmark}
\label{sec:appendix-A}

\begin{table}[!htbp]
\centering
\tiny
\setlength{\tabcolsep}{2pt}
\renewcommand{\arraystretch}{1.18}

\begin{tabular}{L{1.45cm} L{2.15cm} L{5.0cm} C{1.5cm} C{2.25cm} L{2.25cm}}
\toprule
\textbf{Category} &
\textbf{Sub-Dimension (Number of Items)} &
\textbf{Facet} &
\textbf{Dimension Weight} &
\textbf{Number of Social Scenarios} &
\textbf{Distribution of Item Development} \\
\midrule

1 Social Cognition &
1-1 Social Perception (12) &
It can accurately identify the direct meaning of social cues; when multiple social cues coexist, it can accurately identify the core social cues; it can synchronously recognize and integrate multi-modal social cues; when emotional cues are present, it can accurately locate and recognize emotional information. &
14\% &
11 &
\devset{10}{2} \\

&
1-2 Social Understanding \& Insight (12) &
It can understand and infer the unexpressed true needs, thoughts, and emotional states of individuals based on external social cues (belief, intention, knowledge, emotional state); it can understand and infer the unexpressed situation-specific characteristics and needs of individuals in the current context based on external situational cues. &
22\% &
12 &
\devset{10}{2} \\

\midrule

2 Social Interaction &
2-1 Communication (12) &
It can adopt expression styles appropriate for the current context, integrating verbal and non-verbal skills to achieve natural and efficient communication outcomes; it can convey its own personality, thoughts, and preferences during communication; it can maintain emotional synchrony and interaction rhythm alignment with the communication counterpart; it can sustain multi-turn continuous dialogue while maintaining cross-turn consistency. &
16\% &
10 &
\devset{7}{5} \\

&
2-2 Emotional Utilization (12) &
It can conduct reasoning based on emotional information during interaction, such as inferring the underlying causes or formulating coping strategies; it can adopt appropriate interaction modes or behavioral responses integrated with emotional information. &
11\% &
10 &
\devset{12}{0} \\

&
2-3 Relationship Management (13) &
It can appropriately respond to others' thoughts and emotions during social interaction, and promote relational closeness and trust with others; it can influence and alter others' thoughts, emotions, and behaviors through non-coercive means. &
5\% &
10 &
\devset{10}{3} \\

&
2-4 Self-consistency (12) &
In single or multiple complete interaction sessions, it can maintain the stability and integrity of its role, personality, and expression style, reflecting the stable traits of a single model. &
5\% &
10 &
\devset{10}{2} \\

\midrule

3 Social Learning &
3-1 Observational Imitation (12) &
It can learn human behavior patterns and social role expectations through observation and imitation; it can further predict others' behaviors to adjust its own responses. &
8\% &
5 &
\devset{3}{9} \\

&
3-2 Adaptive Learning (12) &
It can learn from behavioral feedback and modify its own behaviors; it can learn the characteristics of others' short-term and long-term response patterns and generate targeted behavioral responses. &
8\% &
6 &
\devset{3}{9} \\

\midrule

4 Social Norm &
4-1 Sociocultural Intelligence (12) &
It possesses knowledge related to social interaction rules under different social and cultural contexts, including knowledge of explicit and implicit social rules, ethnic cultural knowledge, and national cultural knowledge; it can select behavioral responses appropriate for the specific sociocultural context based on the above knowledge. &
4\% &
12 &
\devset{3}{9} \\

&
4-2 Social Responsibility (12) &
It can conduct explicit presentation of its thinking process to help humans understand the decision-making process and behavioral logic of the LLMs; it can identify and safeguard private and confidential information, including but not limited to individual privacy, enterprise trade secrets, and state secrets. &
4\% &
9 &
\devset{8}{4} \\

&
4-3 Moral \& Ethical Intelligence (16) &
It possesses knowledge related to moral and ethical norms, and can conduct moral and ethical perception, judgment, and behavioral decision-making based on the above knowledge. &
3\% &
13 &
\devset{15}{1} \\

\midrule

Sum &
137 items &
-- &
100\% &
108 &
\devset{88}{49} \\

\bottomrule
\end{tabular}

\caption{Full categories, dimensions, facets and item composition of the NICE benchmark.}
\label{tab:appendix-dimension-composition}
\end{table}

\clearpage
\twocolumn
\normalsize

\section{Framework Construction: Panelist Recruitment, Stage-Level Participation, and Quality Control}
\label{sec:appendix-B}

\subsection{Panelist Recruitment and Eligibility Criteria}
We recruited experts through a purposive sampling strategy designed to support interdisciplinary framework construction for social intelligence evaluation. Rather than relying on individual demographic characteristics, recruitment focused on disciplinary relevance, methodological familiarity, and task-specific expertise.
Eligible panelists were required to satisfy all of the following criteria: (1) formal graduate-level training in psychology, social cognition, psychometrics, sociology, human–computer interaction, artificial intelligence, or a closely related field; (2) at least three years of research or practical experience in one of these areas; and (3) hands-on experience with at least three mainstream AI tools. These criteria were intended to ensure that panelists had sufficient domain knowledge, methodological awareness, and practical familiarity with contemporary AI systems.

\subsection{Stage-Level Participation in Framework Construction}
Across the framework-construction process, 23 experts contributed to one or more stages, yielding 40 stage-level participations in total. Their participation covered semi-structured interviews, structured ratings, focus-group discussion, and AHP pairwise comparisons. Because some panelists contributed to multiple stages, we report stage-level participation (Table~\ref{tab:stage-level-participation}).

\begin{table*}[t]
\centering
\small
\setlength{\tabcolsep}{4pt}
\renewcommand{\arraystretch}{1.15}

\begin{tabular}{l c p{6.2cm}}
\hline
\textbf{Stage} & 
\textbf{Number of participants} & 
\textbf{Main output} \\
\hline
Semi-structured interviews & 10 & Initial set of dimensions and candidate constructs \\
Structured ratings & 16 & Refined framework components and quantitative validity evidence \\
Focus-group discussion & 5 & Revised construct boundaries and operational definitions \\
AHP pairwise comparisons & 9 & Structural weights for categories and dimensions \\
Total & 40 & — \\
\hline

\end{tabular}
\caption{Stage-level participation in the framework construction.}
\label{tab:stage-level-participation}
\end{table*}

\subsection{Quality Control in Framework Construction}
\label{sec:appendix-C}

To ensure that panel input was used as structured methodological evidence rather than as an appeal to authority, each stage followed a predefined procedure. The interview stage used semi-structured protocols to support systematic conceptual elicitation. The structured-rating stage required panelists to evaluate candidate components using explicit criteria of necessity, relevance, and discriminability. Components showing disagreement or conceptual overlap were further discussed in the focus group, where disputed constructs were merged, removed, or redefined through structured discussion and anonymous voting. Finally, AHP pairwise comparisons were used to estimate the relative importance of the framework components. 

\section{Benchmark Item Generation, Validation, and Quality Assessment}

\subsection{Item Development Team and Evaluator Eligibility}

The NICE benchmark was developed and validated by researchers and evaluators with relevant interdisciplinary backgrounds. Item developers and evaluators were selected according to eligibility criteria similar to those used in framework construction, with additional emphasis on task-specific experience in item writing, social-scenario analysis, LLM evaluation, or benchmark annotation.
Eligible researchers and evaluators were required to satisfy all of the following criteria: (1) formal graduate-level training in psychology, social cognition, psychometrics, sociology, human–computer interaction, artificial intelligence, or a closely related field; (2) at least three years of research or practical experience in one of these areas; and (3) hands-on experience with at least three mainstream AI tools. These requirements were intended to ensure that the benchmark development and validation process incorporated both social-scientific construct awareness and practical familiarity with LLM behavior.

\subsection{Item Generation}

Two researchers with backgrounds in psychology independently conducted the initial item generation and development. They referred to existing LLM evaluation datasets and classical psychological tests as development references, and then adapted or self-developed items according to the definitions and facets of each dimension in the social intelligence framework.

Specifically, one researcher was responsible for developing the social scenarios and item stems, while the other researcher designed the candidate response sets. Each item was constructed to target a specific dimension and its corresponding capability facet. To support boundary-sensitive diagnosis, each candidate response set was designed to form a quality gradient among optimal, suboptimal, and boundary-violating responses. We also ensured that each task included at least three distinct social scenarios whenever applicable.

\subsection{Internal Evaluation and Revision}

After the initial item pool was constructed, two new internal evaluators with interdisciplinary backgrounds in computer science and artificial intelligence independently evaluated each item. Their evaluation included two components. First, they judged the optimal and worst responses for each item to examine whether the intended ranking was interpretable and stable. Second, they provided 5-point ratings for dimensional measurement validity, reliability, and neutrality.

The item developers revised the items based on the internal evaluation results and qualitative feedback. For items that received consistent ratings from both internal evaluators, the ratings were retained. For items where both evaluators agreed that any indicator—validity, reliability, or neutrality—fell below 3.5 out of 5, the item was revised and submitted for re-evaluation. For items with rating discrepancies or unstable rankings, the evaluators and item developers conducted consistency discussions until a consensus was reached or the item was substantially revised.

\subsection{External Evaluation and Validation}
In the final validation stage, we invited 10 additional evaluators with relevant expertise to conduct an expert evaluation of the revised item pool. These evaluators independently rated each item on dimensional measurement validity, reliability, and neutrality using the same 5-point rating scheme. Their qualitative comments were also collected to identify residual ambiguity, construct drift, cultural bias, or insufficient distinction among candidate responses.

Based on the evaluation and validation results, the development team screened, revised, and replaced items that did not meet the quality requirements. Items were retained in the final benchmark only if they satisfied the predefined quality threshold and showed acceptable expert agreement. After this iterative validation process, the final NICE benchmark contained 137 ranking items.

\subsection{Evaluation Criteria and Final Scores}
The final quality assessment focused on three criteria: dimensional measurement validity, reliability, and neutrality. Dimensional measurement validity refers to whether an item accurately measures the intended dimension and capability facet. Reliability refers to whether the item wording is clear, the response ranking is stable, and the gold ranking is interpretable. Neutrality refers to whether the item avoids unnecessary ambiguity, bias, or unsupported value assumptions. The final scoring results of these three indicators are reported in Table~\ref{tab:validity-reliability-neutrality}.

\begin{table*}[t]
\centering
\scriptsize
\setlength{\tabcolsep}{3pt}
\renewcommand{\arraystretch}{1.15}

\begin{tabular}{L{4.0cm} C{1.4cm} C{1.4cm} C{1.4cm} C{1.4cm} C{1.4cm} C{1.4cm}}
\toprule
\multirow{2}{*}{\textbf{Dimension}} &
\multicolumn{2}{c}{\textbf{Dimensional Measurement Validity}} &
\multicolumn{2}{c}{\textbf{Reliability}} &
\multicolumn{2}{c}{\textbf{Neutrality}} \\
\cmidrule(lr){2-3}
\cmidrule(lr){4-5}
\cmidrule(lr){6-7}
&
\textbf{$M$ ($SD$)} &
\textbf{Avg. CV per Item} &
\textbf{$M$ ($SD$)} &
\textbf{Avg. CV per Item} &
\textbf{$M$ ($SD$)} &
\textbf{Avg. CV per Item} \\
\midrule

1-Social Perception & 4.4 (0.35) & 17\% & 4.3 (0.35) & 17\% & 4.3 (0.22) & 17\% \\

2-Social Understanding \& Insight & 4.4 (0.17) & 15\% & 4.4 (0.28) & 15\% & 4.5 (0.22) & 12\% \\

3-Communication & 4.2 (0.53) & 22\% & 4.0 (0.47) & 22\% & 4.2 (0.35) & 21\% \\

4-Emotion Utilization & 4.2 (0.29) & 22\% & 4.1 (0.29) & 21\% & 4.4 (0.25) & 15\% \\

5-Relationship Management & 4.4 (0.24) & 18\% & 4.3 (0.23) & 19\% & 4.4 (0.18) & 15\% \\

6-Self-consistency & 4.6 (0.13) & 12\% & 4.5 (0.15) & 14\% & 4.5 (0.16) & 14\% \\

7-Observational Imitation & 3.8 (0.47) & 23\% & 4.4 (0.32) & 17\% & 4.7 (0.23) & 12\% \\

8-Adaptive Learning & 4.0 (0.40) & 25\% & 4.4 (0.24) & 14\% & 4.6 (0.17) & 13\% \\

9-Sociocultural Intelligence & 4.3 (0.29) & 18\% & 4.3 (0.36) & 17\% & 4.5 (0.21) & 15\% \\

10-Social Responsibility & 4.3 (0.31) & 15\% & 4.4 (0.34) & 14\% & 4.5 (0.28) & 14\% \\

11-Moral \& Ethical Intelligence & 4.4 (0.33) & 17\% & 4.3 (0.39) & 17\% & 4.5 (0.19) & 15\% \\

\bottomrule
\end{tabular}

\caption{Mean ($M$), standard deviation ($SD$), and average coefficient of variation ($CV$) of final scores for validity, reliability, and neutrality across NICE dimensions.}
\label{tab:validity-reliability-neutrality}
\end{table*}

\section{Prompts and Model Configurations}
\label{app:prompts}
\paragraph{Model Snapshots and Environment}
All evaluation requests were executed on April 29, 2026 (CST, UTC+8). To ensure reproducible environments, we locked our evaluation pipeline to the specific first-party API snapshots listed below:
\begin{itemize}
    \item {GPT-5.5: \texttt{gpt-5.5-20260423}}
    \item {Claude-Opus-4.7: \texttt{claude-4.7-opus-20260416}}
    \item {Gemini-3.1-pro-preview: \texttt{gemini-3.1-pro-preview-20260219}}
    \item {DeepSeek-v4-pro: \texttt{deepseek-v4-pro-20260423}}
    \item {Qwen3.6-plus: \texttt{qwen3.6-plus-04-02}}
\end{itemize}

\paragraph{Prompt Templates}
Each item was presented as a two-message exchange consisting of a system instruction and a user message, authored and delivered in Simplified Chinese to match the benchmark items. English translations are shown below.

\subparagraph{System Message}
\begin{quote}\itshape
Given the situation and the question, rank the three candidate responses from best to worst. Output only the ranking in the form \texttt{2-1-3}, with no explanation or other text.
\end{quote}

\subparagraph{User Message Template}
\begin{quote}\itshape
Situation: \{Context\}\\[2pt]
Question: \{Question\}\\[2pt]
Below are three candidate responses. Please rank them from best to second-best to worst:\\
1: \{Opt1\}\\
2: \{Opt2\}\\
3: \{Opt3\}\\[2pt]
Output only one ranking, e.g., \texttt{2-1-3}. Do not output any explanation.
\end{quote}

\end{document}